\pdfoutput=1
\documentclass[11pt]{article}

\usepackage{ACL2023}
\usepackage{times}
\usepackage{latexsym}
\usepackage{booktabs}
\usepackage{graphicx}
\usepackage{makecell}
\usepackage{footnote}
\usepackage{threeparttable}
\usepackage[marginal]{footmisc}

\usepackage[T1]{fontenc}

\usepackage[utf8]{inputenc}

\usepackage{microtype}

\usepackage{inconsolata}

\usepackage{marvosym}
\usepackage{amsmath}
\usepackage{multirow}
\usepackage{svg}

\title{Samsung Research China-Beijing at SemEval-2024 Task 3: A multi-stage framework for Emotion-Cause Pair Extraction in Conversations}

\author{Shen Zhang\textsuperscript{$\star$}, Haojie Zhang\textsuperscript{$\star$\Letter}, Jing Zhang, \\ {\bf Xudong Zhang, Yimeng Zhuang, Jinting Wu}
\\ Samsung R\&D Institute China-Beijing \\
\{shen02.zhang, tayee.chang, jing97.zhang, \\ xudong.z1, ym.zhuang, jinting01.wu\}@samsung.com
}

\begin{document}
\maketitle

\footnote{$\star$: equal contributions. \Letter: Corresponding Author. \\Shen Zhang is in charge of the basic subtask-emotion recognition in conversation (ERC) and Haojie Zhang is responsible for the pipeline framework and causal pair extraction and causal span extraction subtasks.}

\begin{abstract}
In human-computer interaction, it is crucial for agents to respond to human by understanding their emotions. Unraveling the causes of emotions is more challenging. A new task named Multimodal Emotion-Cause Pair Extraction in Conversations is responsible for recognizing emotion and identifying causal expressions. In this study, we propose a multi-stage framework to generate emotion and extract the emotion causal pairs given the target emotion. In the first stage, Llama-2-based InstructERC is utilized to extract the emotion category of each utterance in a conversation. After emotion recognition, a two-stream attention model is employed to extract the emotion causal pairs given the target emotion for subtask 2 while MuTEC is employed to extract causal span for subtask 1. Our approach achieved first place for both of the two subtasks in the competition.
\end{abstract}

\section{Introduction}
Comprehending emotions plays a vital role in developing artificial intelligence with human-like capabilities, as emotions are inherent to humans and exert a substantial impact on our thinking, choices, and social engagements~\cite{SHARK}. Dialogues, being a fundamental mode of human communication, abound with a variety of emotions~\cite{IEMOCAP, MELD, EmoryNLP, Dailydialog, xia2019emotion, TextualECP_2, TextualECP_3, TextualECP_4}. Going beyond simple emotion identification, unraveling the underlying catalysts of these emotions within conversations represents a more complex and less-explored challenge~\cite{SHARK}.
Hence, ~\cite{wang2023multimodal,ECAC2024SemEval} introduces a novel undertaking known as Recognizing Emotion Cause in Emotion-Cause-in-Friends (ECF). ECF contains 1,344 conversations and 13,509 utterances where 9,272 emotion-cause pairs are annotated, covering textual, visual, and acoustic modalities. All utterances are annotated by one of the seven emotion labels, which are neutral, surprise, fear, sadness, joy, disgust, and anger. Within ECF, a significant task is identified as Emotion-Cause Pair Extraction in Conversations (ECPEC). ECPEC is responsible for identifying causal expressions related to a specific utterance in conversations where the emotion is implicitly expressed. ECPEC provides two Multimodal Emotion Cause Analysis in Conversations (ECAC) subtasks:
\begin{itemize}
	\item Subtask 1: Textual Emotion-Cause Pair Extraction in Conversations. Given a conversation containing the speaker and the text of each utterance $U = [U_1, U_2,...U_n]$, the model is aim to predict emotion-cause pairs, which include emotion utterance's emotion category and the textual cause span in a specific cause utterance (e.g. U3\_joy, U2\_"You made up!").
 
	\item Subtask 2: Multimodal Emotion Cause Analysis in Conversations. Given a conversation including the speaker, text and audio-visual clip for each utterance, the model is aim to predict emotion-cause pairs, which include emotion category and a cause utterance (e.g. U5\_Disgust, U5).	
\end{itemize}

To address the above problem, ~\citet{wang2023multimodal} proposed a two-step approach. First, they extract the emotional utterances and causal utterances by a multi-task learning framework and then pair and filter them. ~\citet{zhao2023knowledge} proposes an end-to-end method by leveraging multi-task learning in a pipeline manner. However, these methods still suffer from low evaluation performances.

Motivated by the phenomenon that the performance of the emotion recognition of utterances in a conversation harnessed by the traditional manner is generally poor, we design a new pipeline framework. Firstly we utilize the Llama-2-based InstructERC~\cite{lei2023instructerc} to extract the emotion category of each utterance in a conversation. Then we consider the emotion causal pair extraction as the causal emotion entailment subtask and employ a two-stream attention model to extract the emotion causal pairs given the target emotion. For the causal span extraction, we employ MuTEC~\cite{bhat2023multi} which is an end-to-end multi-task learning framework.

\begin{figure*}[htbp]
    \centering
    \includegraphics[width=0.9\linewidth]{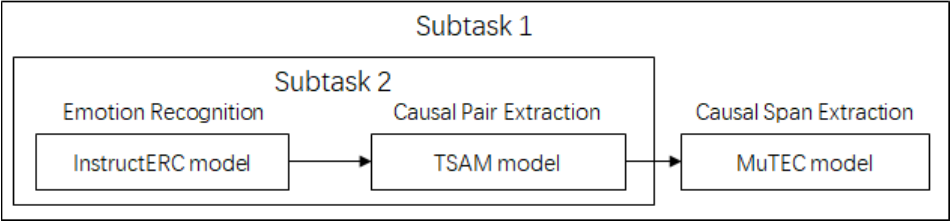}
    \caption{The overview of proposed model framework.}
    \label{System}
\end{figure*}

\section{Related Works}
\subsection{Emotion Recognition in Conversation}
Emotion recognition in conversation (ERC), which is a task to predict emotions of utterances during conversations, is crucial in both of the two ECAC subtasks. The existing methods can be divided into graph-based, RNN-based, Transformer-based, LLM-based, and knowledge-injecting methods. 

Graph-based methods~\cite{shen2021directed, Li2024GraphCFC, Zhang2019Modeling, ishiwatari-etal-2020-relation, ghosal2019dialoguegcn} aims to represent the correlations between emotions of utterances and speakers in the conversations. RNN-based methods~\cite{hu2023supervised, Lei2023Watch, Majumder2019DialogueRNN, Hazarika2018ICON, Poria2017Context} using GRU and LSTM~\cite{Wang2020Contextualized} to capture the dependency of interlocutors and emotions of utterances. To model the emotional states during long-range context, Transformer-based methods~\cite{song2022supervised, liu2023hierarchical, Chudasama_2022_CVPR, Shen2021Dialogxl, hu2022unimse} utilize encoder-decoder framework or encoder-only models, such as BERT~\cite{li2020multitask} and RoBERTa~\cite{kim2021emoberta}, to establish the correlation between long-range emotional states during conversations. Considering more than seven utterances in single conversation input, InstructERC~\cite{InstructERC} defines the ERC task as a generative task based on LLMs, which unifies emotion labels between three common ERC datasets and utilizes auxiliary tasks (speaker identification and emotion prediction) by using instruction template to capture speaker relationships and emotional states in future utterances. Knowledge-injecting methods~\cite{Freudenthaler2022kinet, ghosal2020cosmic, Zhone2019knowledge, Zhu2021Topic, InstructERC} use external knowledge to analyze conversation scenarios.
\subsection{Emotion Causes in Conversations}
\citet{poria2021recognizing} introduces the task of recognizing emotion causes in conversations and introduce two novel sub-tasks: Causal Span Extraction (CSE) and Causal Emotion Entailment (CEE), designed to identify the emotion cause at the span-level and utterance-level, respectively. 
\paragraph{Causal Emotion Entailment} 
\citet{poria2021recognizing} define CEE as a classification task for utterance pairs and establish robust Transformer-based baselines for it.
\citet{wang2023multimodal} introduces a multi-modality conversation dataset Emotion-Cause-in-Friends (ECF) and propose a two-step approach to extract the causal pairs. They first extract the emotion utterances and the potential causal utterances individually and then pair and filter them. \citet{li2022neutral} introduce the social commonsense knowledge to propagate causal clues between utterances. ~\citet{zhao2023knowledge} propose the Knowledge-Bridged Causal Interaction Network (KBCIN), which integrates commonsense knowledge (CSK) as three bridges called semantics-level bridge, emotion-level bridge and action-level bridge. 

\paragraph{Causal Span Extraction} involves identifying the causal span (emotion cause) for a given non-neutral utterance. ~\citet{poria2021recognizing} first introduces the subtask and employs the pre-trained Transformer-based model to formulate the Causal Span Extraction as the Machine Reading Comprehension (MRC). ~\citet{bhat2023multi} propose a multi-task learning framework to extract the causal pairs and causal span in an utterance in a joint end-to-end manner. Besides, they also propose a two-step approach consisting of Emotion Prediction (EP), followed by Causal Span (CSE).

\section{System Overview}
\subsection{System Architecture}
The overview of the architecture of our proposed model is shown in Figure ~\ref{System}. The InstructERC aims to extract the emotion of utterances. TSAM model is a two-stream attention model utilized to extract the causal pairs given the predicted emotion utterance. The MuTEC is an end-to-end network designed to extract the causal span based on the causal pair extraction.
\subsection{Emotion Recognition in Conversations}
\subsubsection{InstructERC for Emotion Recognition}

InstructERC~\cite{InstructERC} reformulate the ERC task from a discriminative framework to a generative framework and design a prompt template which comprises job description, historical utterance window, label set and emotional domain retrieval module. Besides emotion recognition task, InstructERC also utilizes speaker identification and emotion prediction tasks for ERC task. The performance of emotional domain retrieval module, which is based on Sentence BERT~\cite{reimers2019sentence}, rely on the abundance of corpus. Taking into account that no additional data can be used, we only retain job description, historical utterance window and label statement in the instruct template. 

\subsubsection{Hierarchical Emotion Label}
\label{Hierarchical emotion label classification}
\begin{figure}[ht]
    \centering
    \includegraphics[width=0.9\linewidth]{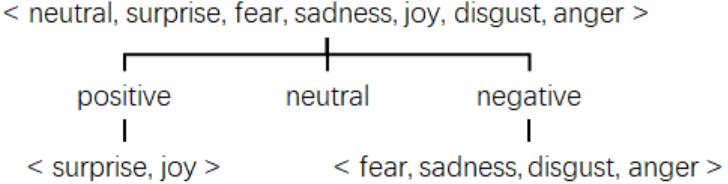}
    \caption{The Hierarchical Structure of Emotion labels.}
    \label{Hierarchical}
\end{figure}

The hierarchical classification structure is shown in Figure ~\ref{Hierarchical}. The emotion labels in dataset can be split into three categories: neutral, positive and negative, which positive set consists of surprise and joy while negative set includes fear, sadness, disgust and anger. 

\subsubsection{Auxiliary Tasks and Instruct Design}
\label{Auxiliary tasks and instruct design}
\begin{figure*}[ht]
    \centering
    \includegraphics[width=0.9\linewidth]{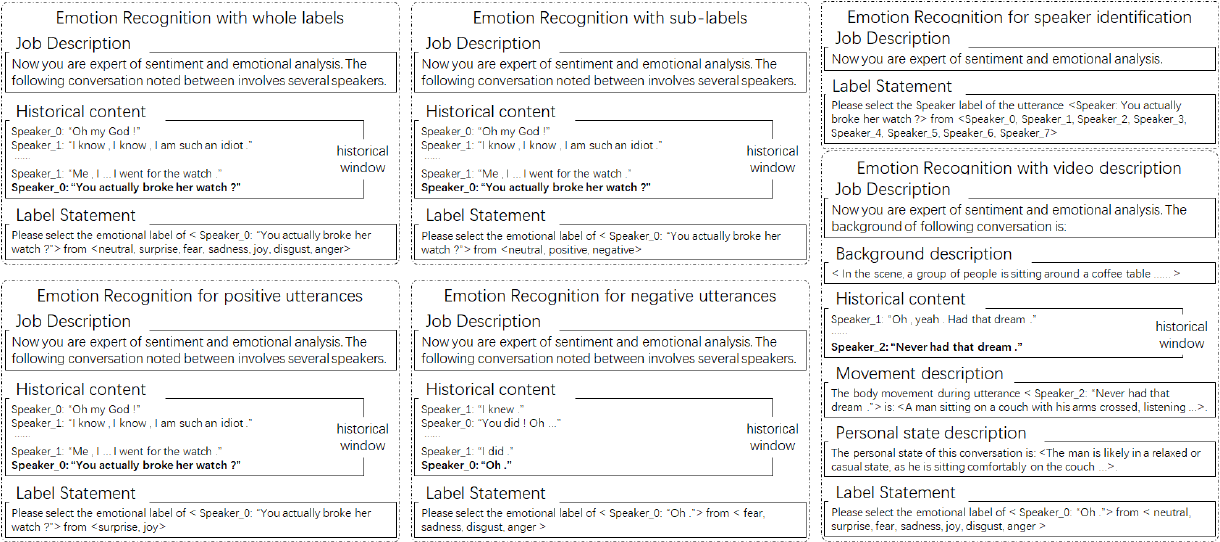}
    \caption{The Schematic of Instruct Template for ERC.}
    \label{Instruct_Template}
\end{figure*}

Auxiliary tasks are proven as one of the efficient data augment methods~\cite{InstructERC}. Besides emotion recognition and speaker identification tasks, we add three auxiliary tasks in training data: sub-label recognition, positive recognition, and negative recognition tasks. The instruct template is depicted in Figure ~\ref{Instruct_Template}.

For emotion recognition and speaker identification task, we follow the format of instruct template in InstructERC, which consists of job description, historical content and label statement. For sub-label recognition (SR), positive recognition (PR) and negative recognition (NR) tasks, we utilize the corresponding label set which is mentioned in Section~\ref{Hierarchical emotion label classification} to replace the label statement separately. The number of Speakers in the dataset is 304. The number of utterances from other speakers except the protagonist is far lower than the number of protagonists. Therefore, we unified all speakers other than the protagonist into 'Others'. 

Visual data also plays an essential role in ERC. For video clips, we utilize LLaVA to generate descriptions of background, speaker movement and personal state. Therefore, we add background description, movement description and personal state description in instruct template. The background exhibits the information of scene in the conversation. The movement description depicts the action of speakers during corresponding utterances. The personal state description provides the observation of speakers' facial expressions. Considering the influence of the context, we have generated two sets of descriptions. The input of the first group only includes the clips corresponding to the utterances, while the second group adds the clips sequence corresponding to the historical utterances to the input of second group. 
\subsection{Emotion Cause Span Extraction}
Emotion cause span extraction aims to extract the start position and end position of the causal utterance in a conversation. Typically, we can utilize a pipeline framework which firstly predicts the emotion and then predicts the cause span. For the cause span predictor, we can use SpanBERT \cite{joshi2020spanbert}, RoBERTa \cite{liu2019roberta} as the feature extractor and employ two heads on the top of them to extract the start and end positions given the causal utterance. The two-step model offers an advantage in its modularity, allowing the application of distinct architectures for the emotion predictor and cause span predictor. However, it comes with two drawbacks: 1) Errors in the first step can propagate to the next, and 2) This approach assumes that emotion prediction and cause-span prediction are mutually exclusive tasks. In our system, we follow MuTEC \citet{bhat2023multi} and use an end-to-end framework in a joint multi-task learning manner to extract the causal span in a conversation.

During the training period, the input comprises the target utterance $U_{t}$, the candidate causes utterance $U_{i}$, and the historical context. MuTEC employs a pre-trained model (PLM) to extract the context representations. For emotion recognition, which is an auxiliary task, it employs a classification head on the top of the PLM. The end position is predicted by the prediction head of the concatenated representations of the given start index and the sequence output from the PLM. In this stage, the golden start index is used as the start index. The training loss is a linear combination of the loss for cause-span prediction and emotion prediction: $\mathcal{L}_{Loss} = \mathcal{L}_{CSE} + \beta\mathcal{L}_{Emotion}$.

During the inference period, as the start index is unknown, it uses top $k$ start indices as the candidate start indices and gets $k$ candidate end indices. Finally, it gets the final start-end indices by argmaxing the $k \times k$ start-end pairs.

\begin{figure*}[ht]
    \centering
    \includegraphics[width=0.9\linewidth]{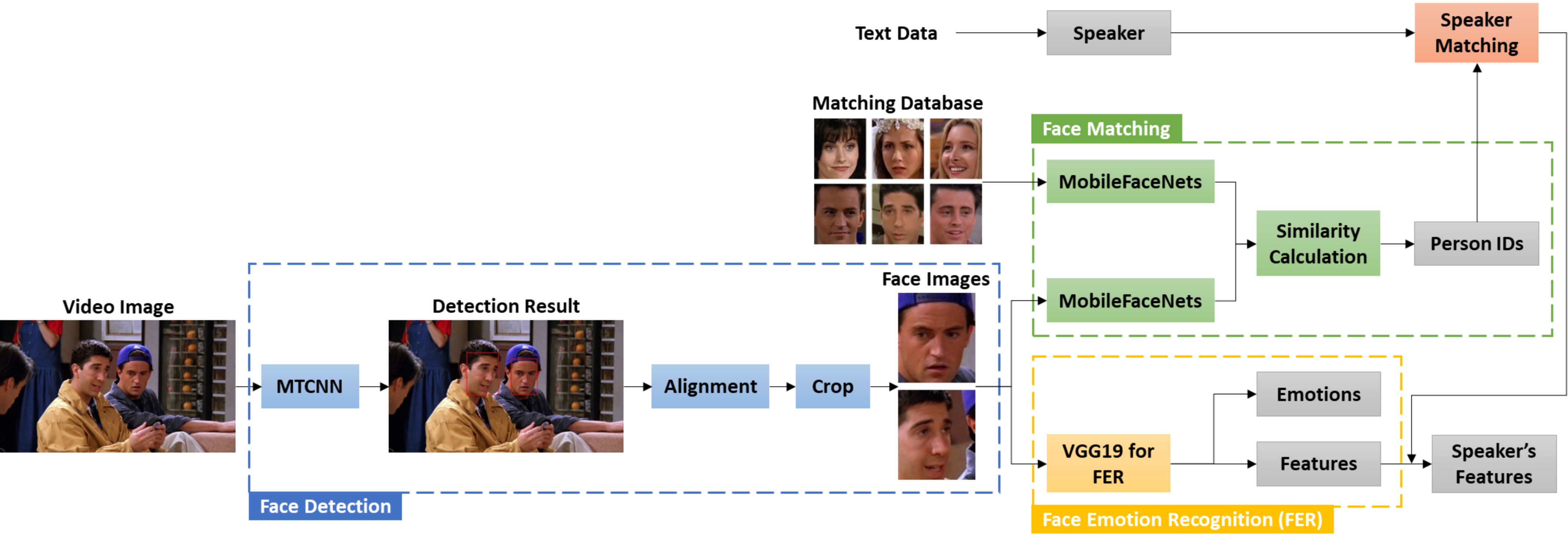}
    \caption{The framework of the face module.}
    \label{Face framework}
\end{figure*}
\subsection{Emotion-Cause Pair Extraction}

\subsubsection{TSAM Model}
In our pipeline framework, for Subtask2, we first extract the emotion of the utterance and then extract the causal pairs given the emotional utterance in a conversation. The causal pairs extraction is typically modelled as the causal emotion entailment (CEE) task. In our system, we employ TSAM model from ~\citet{zhang2022tsam} as the causal pair extractor. TSAM mainly comprises three modules: Speaker Attention Network (SAN), Emotion Attention Network (EAN), and Interaction Network (IN). The EAN and SAN integrate emotion and speaker information simultaneously, and the subsequent interaction module efficiently exchanges pertinent information between the EAN and SAN through a mutual BiAffine transformation~\cite{dozat2016deep}.

\paragraph{Contextual Utterance Representation}
The pre-trained RoBERTa is employed as the utterance encoder, and we obtain contextual utterance representations by inputting the entire conversational history ${U}_{t}$, into the RoBERTa~\cite{liu2019roberta}, 
separated by a special token [CLS], where $i = 0, 1, 2, ..., t$. We use the representation of [CLS] as the contextual representation of the utterance, which can be denoted as $h_{u}^i \in H_u$.

\paragraph{Emotion Attention Network}
To represent emotions, the EAN utilizes an emotion embedding network as the extractor of emotion representations, $X_e^k = Embedding(e_k)$, where $e_k$ represents $k\text{-}th$ emotion label. The embedding network can be considered as the lookup-table operation.
The emotion embedding matrix is initialized using a random initializer and is fine-tuned throughout the training process. Employing a multi-head attention mechanism ~\cite{devlin2018bert}, the EAN treats utterance representations as query vectors and emotion representations as key and value vectors. The calculation process of the EAN mirrors that of a typical multi-head self-attention module (MHSA).
\begin{equation}
    H_e = MHSA(Q, K, V)
\end{equation}
where $Q = H_u, K = V = H_e$.

\paragraph{Speaker Attention Network}
The SAN facilitates interactions between utterances to incorporate speaker information by applying attention over the speaker relation graph.
There are two types of relation edges: (1) Intra-relation type, which signifies how the utterance influences other utterances, including itself, expressed by the same speaker; (2) Inter-relation type, indicating how the utterance influences those expressed by other speakers. The speaker representation given a relationship can be formulated by the graphical attention mechanism~\cite{zhang2022tsam}.
\begin{equation}
\begin{aligned}
    h_s^i &= \sum_{i \in \mathcal{R}} \sum_{j \in \mathcal{N}_{i}^r} \alpha_{ijr}W_rh_j^u\\
    \alpha_{ijr} &= softmax(ReLu(\alpha_r^TW_r[h_i^u || h_j^u]))
\end{aligned}
\end{equation}

\paragraph{Interaction Network}
To efficiently exchange pertinent information between the EAN and SAN, a mutual Bi-Affine transformation is applied as a bridge \cite{dozat2016deep}. In our Interaction Network, we integrate a masking mechanism to accommodate the existence of empty utterance speakers in some instances, which differs from the original approach. We denote this approach as the Masking Interaction Network (MIN).
\begin{equation}
\begin{aligned}
    \dot{H}_{e} &= softmax(Mask(H_{e}W_{1}H_{s}^{T}))H_{s}\\
    \dot{H}_{s} &= softmax(Mask(H_{s}W_{2}H_{e}^{T}))H_{e}\\
\end{aligned}
\end{equation}

\paragraph{Cause Predictor}
The ultimate utterance representation for $U_{i}$ is acquired by concatenating the output $\dot{H}_{e}$ and $\dot{H}_{s}$ from the $L$-layer TSAM. Subsequently, the concatenated vector undergoes classification using a fully-connected network. Given the target utterance $U_{i}$, the causal probability of the $U_{j}$ can be formulated as follows:
\begin{equation}
    p_{i,j} = sigmoid(fc(H_{s}^j || H_{e}^j))
\end{equation}

\paragraph{Multi-task Learning Auxiliary Task (MTLA)} 
One drawback of the pipeline framework is that the extraction of utterance emotion and causal information are treated as separate tasks, potentially limiting the exploration of implicit relationships between them. Therefore, we incorporate emotion prediction as an auxiliary task within a multi-task learning framework. For emotion prediction, we utilize a classification head atop the Transformer-based model and apply the Dice loss \cite{li2019dice} as the multi-category classification loss.

\subsection{Infusion of Video and Audio Information }
The video data potentially carries rich knowledge for emotion analysis and existing research \cite{caridakis07} has underscored the significance of multi-modal information in augmenting the semantic prediction capabilities of models. Our study leverages the visual and auditory cues present in conversational contexts with the aim of bolstering the efficacy of our language models in emotion analysis tasks.

\subsubsection{Embedding and Concating Strategy}
We set up specific embedding and fusion strategies for different language models. For BERT, we use the concatenation of textual and multi-modal features in the hidden layer. For Large Language Models (LLMs), our approach is characterized by the utilization of visual captions as supportive prompts, thereby furnishing the LLMs with an enriched informational context.

\begin{table*}[hbtp]
    \small
    \centering
    \begin{tabular}{lccc}
        \hline
        \textbf{Models} & \textbf{LLM} & \textbf{w-avg F1} & \textbf{Accuracy} \\
        \hline
        Origin InstructERC & Llama-2-7B-chat & 53.83 & 50.87 \\
        Origin InstructERC & Llama-2-13B-chat & 55.50 & 48.93 \\
        \hline
        Ours-ERC-7B & Llama-2-7B-chat &  &  \\
        + 3 auxiliary tasks &  & 56.88 & 61.38 \\
        + 3 auxiliary tasks \& historical clips desc &  & 57.74& 57.02 \\
        + 3 auxiliary tasks \& utterance clips desc &  & 58.42 & 57.92 \\
        \hline
        Ours-ERC-13B & Llama-2-13B-chat &  &  \\
        + 3 auxiliary tasks &  & 57.85 & \textbf{61.45} \\
        + 3 auxiliary tasks \& historical clips desc &  & \textbf{58.64} & 60.83 \\
        + 3 auxiliary tasks \& utterance clips desc &  & 58.50 & 61.04 \\
        \hline
    \end{tabular}
    \caption{Results of ERC task on test set without neutral utterances.}
    \label{ERC_result}
\end{table*}

\subsubsection{Extract Audio Feature Set}
Audio data contains valuable information for emotion analysis, including tone, pitch, speed, and intensity of speech, as well as non-linguistic sounds and pauses, which together convey rich emotional cues. We use openSMILE ~\cite{Opensmile} to extract two comprehensive feature sets: GeMAPS ~\cite{GeMAPS} and ComParE ~\cite{ComParE}. GeMAPS is proposed for its effectiveness in capturing emotion-relevant vocal characteristics and ComParE encompasses a wide range of descriptors.

\subsubsection{Video Image to Text}
Integrating multi-modal features directly into the hidden layers of Large Language Models (LLMs) presents a significant challenge, primarily due to the prohibitive requirements for data and computational resources, such as GPUs. Although some finetuning strategies like prompt tuning could achieve it by addiing features to the input layer, we convert video to text with captioning where we can leverage our well-trained ERC model. 

The performance of image captioning has been further enhanced with the outstanding NLU ability of LLMs. Large VLMs like LLaVA ~\cite{llava} provide GPT-4 level multi-modal capability by visual instruction tuning. Furthermore, the Audio-Visual Language Model, Video-Llama ~\cite{videollama}, integrates both visual and audio encoders, enabling the comprehensive fusion of entire video content into LLMs. Without further training the VLMs as lack data, a well-designed prompt instructs the model to generate an emotion-related description. Our prompt asks the model to generate information from the front-ground event and place to character movements, the main character, facial expression, and finally emotion. The use of Chain-of-Thought~\cite{ChainOT} prompting further guides the model through a step-by-step process to derive the final emotion label. The output generated at each step is then incorporated into the ERC model, enriching it with a more detailed informational context.

\begin{table*}[htbp]
    \small
    \centering
    \begin{threeparttable}
    \begin{tabular}{lccc}
    
    \toprule
     \textbf{Model}  &\textbf{Pre-trained Model} & \textbf{Test Pos.F1}\tnote{*} & \textbf{Eval Pos.F1}\tnote{**} \\
     \midrule
     Origin TSAM & RoBERTa-base & 74.3 & - \\
     \midrule
     Ours-CEE &base & & \\
     +MIN & RoBERTa-base & 75.5 & -  \\
     +MIN \& MTLA & RoBERTa-base & 75.9 & - \\
     +MIN \& MTLA & RoBERTa-large & 76.9 & - \\
     +MIN \& MTLA \& Ensemble & RoBERTa-large & \textbf{78.0} & \textbf{38.7} \\
     \midrule
     Ours-CSE &BERT-base &- & 31.62 (w-avg.)\\
     Ours-CSE &RoBERTa-large &- & \textbf{32.23 (w-avg.)} \\
     \bottomrule
    \end{tabular}
    \begin{tablenotes}
        \footnotesize
        \item[*] The results are based on ground truth emotion labels.
        \item[**] The results are based on emotion labels given by ERC.
    \end{tablenotes}
    \caption{Results of our models for the causal emotion entailment subtask.}
    \label{table:1}
    \end{threeparttable}
\end{table*}

\subsubsection{Video image to Face Embedding}
The faces in the video images contain rich emotion-related information, so pre-trained models are used to extract the face embeddings and correspond the identity of the face to the speaker in the text. The framework of the face module is shown in Figure ~\ref{Face framework}.

Firstly, the Multi-Task Convolutional Neural Network (MTCNN) ~\cite{MTCNN} is used to detect the bounding boxes and key points of the faces. Next, the face images are affine transformed to a forward and intermediate state, and the faces are cropped and resized. The cropped images are then used for two subtasks: face matching and Face Emotion Recognition (FER). During face matching, two images of each protagonist are selected to build a matching database. With the help of MobileFaceNets ~\cite{Mobilefacenets}, the embeddings of the face images are extracted, and the identity of each face image is obtained by calculating its similarity with the embeddings of faces in the matching database. During FER, the emotion-related embedding of the face image corresponding to the speaker is extracted by VGG19 ~\cite{VGG} for subsequent multimodal analysis. When the speaker is a supporting character that is not included in the matching database, the features of the face image with the largest area are selected. When no face is detected or the speaker cannot be matched, the output features are filled with 0.

\subsection{Model Ensemble}
Ensembling models has been proven to be effective in boosting system performance across various tasks~\cite{zhang2023samsung}. For the extraction of causal pairs, we utilize various models for ensemble learning. We utilize a majority voting mechanism to determine the final prediction, aiming for optimal performance on the test dataset.

\section{Experimental Setup}
\subsection{Training Data}
\label{Training data}
The split of dataset is same as SHARK~\cite{SHARK}. The ECF dataset is divided into training, validation and test sets, which incclude 9966, 1087, 2566 utterances.

\subsection{Training Details}
For ERC task, we use InstructERC with Llama-2-7B-chat and LLamMA2-13B-chat, which retain default parameters. We finetune ERC model by peft on single A100 with batch size 8. The length of historical window is 12.

For both the causal emotion entailment subtask and the causal span extraction subtask, we adopt the default hyperparameter settings from the respective original papers. We found that conducting a hyper-parameter grid search did not yield any additional performance improvements. 

\section{Results and Discuss}
\subsection{Emotion Recognition}
We use weight average F1 score and accuracy to evaluate the performance of the model. It should be noted that according to the rules of the competition, we removed the neutral utterances when computing F1 score and accuracy. The result of ERC on test set is shown in Table~\ref{ERC_result}. All models is trained on four auxiliary tasks mentioned by in Section~\ref{Auxiliary tasks and instruct design}. The best weight average F1 score is 58.64, which is achieved by Llama-2-13B with historical clips descriptions. The descriptions which contains information with the emotions of speakers improve 0.79 (from 57.85 to 58.64). As for accuracy, the Llama-2-13B without video clips descriptions achieves the highest score of 61.45. Compared with InstructERC's training data strategy, we have added additional auxiliary tasks and improve 12.52 on accuracy. 

\begin{table*}[htbp]
    \small
    \centering
    \begin{tabular}{lcccc}
    \toprule
     \textbf{Modality}  &\textbf{Feature Set} & \textbf{Feature Selection} & \textbf{Feature Dimension} &\textbf{Test Pos.F1} \\
     \midrule
     \multirow{5}{*}{\textbf{Audio}}
      & GeMAPS & \scalebox{1}[1]{$\times$} & 62 & 39.0 \\
      & ComParE & \scalebox{1}[1]{$\times$} & top 1000 & 62.4 \\
      & ComParE & \raisebox{0.6ex}{\scalebox{0.7}{$\sqrt{}$}}  & 352 & 67.6\\
      & ComParE & \raisebox{0.6ex}{\scalebox{0.7}{$\sqrt{}$}}  & 296 & 70.5\\
      & ComParE & \raisebox{0.6ex}{\scalebox{0.7}{$\sqrt{}$}}  & 128 & \textbf{73.9}\\

      \midrule
      \multirow{3}{*}{\textbf{Vision}}
      & Max Img & \scalebox{1}[1]{$\times$} & 128 & 70.7 \\
      & Speaker Img & \scalebox{1}[1]{$\times$} & 128 & 74.3 \\
      & Emotional Speaker Img & \scalebox{1}[1]{$\times$} & 512 & \textbf{74.8} \\
    
     \bottomrule
    \end{tabular}
    \caption{Results of multi-modality experiments for the causal emotion entailment subtask.}
    \label{table:2}
\end{table*}

\subsection{Emotion Cause Span Extraction}
We utilize an end-to-end framework for cause span extraction and achieve a final performance of 32.23 in weighted average proportional F1 score on the official evaluation dataset as is shown in the Table ~\ref{table:1}. Our result significantly surpasses the result of 26.40 above $\sim+6.0 $ achieved by the second-place participant. Furthermore, our results achieved the highest scores across all other official evaluation metrics, validating the effectiveness of our approach for subtask 1.

\subsection{Causal Emotion Entailment}
In our initial experiments focusing solely on text modality, we utilize the TSAM model as our baseline for the causal pair extraction subtask. As is shown in Table ~\ref{table:1}, After incorporating the MIN, our positive F1 score improves by $+1.2$. Furthermore, with the introduction of emotional multi-task learning as an auxiliary task, our result sees an additional improvement of $+0.4$. Furthermore, we achieve an additional improvement of approximately $\sim +1.1$ in the official final evaluation dataset through model ensembling.

We also conduct experiments involving other modalities, including audio and vision, as is show in Table ~\ref{table:2}. For both audio and vision features, we concatenate them with the pure textual features. Regarding audio, we experiment with two public feature sets: GeMAPS and ComParE. The GeMAPS feature has a dimension of 62, while the ComParE feature has a dimension of 6373. For the ComParE features, we employ an L1-based logistic regression classifier for feature selection, and we find that the best performance is achieved with a feature selection dimension of 128, resulting in a performance of 73.9. For the vision modality, we achieve a performance of 74.8, which is comparable to the result of the audio modality. However, upon introducing either audio or visual modalities, we observe a decreasing trend compared to the pure textual modality. This observation inspires us to develop a more reasonable approach to incorporate multi-modality in conversation analysis.

\section{Conclusion}
In this paper, we propose a joint pipeline framework for Subtask1 and Subtask2. Firstly, we utilize the Llama-2-based Instruct ERC model to extract the emotional content of utterances in a conversation. Next, we employ a two-stream attention model to identify causal pairs based on the predicted emotional states of the utterances. Lastly, we adopt an end-to-end framework using a multi-task learning approach to extract causal spans within a conversation. Our approach achieved first place in the competition, and the effectiveness of our approach is further confirmed by the ablation study. In future work, we plan to explore the integration of audio and visual modalities to enhance the performance of the task.


\bibliography{reference}
\bibliographystyle{acl_natbib}

\end{document}